\def\year{2020}\relax
\documentclass[letterpaper]{article} 
\usepackage{aaai20}  
\usepackage{times}  
\usepackage{helvet} 
\usepackage{courier}  
\usepackage[hyphens]{url}  
\usepackage{graphicx} 
\usepackage{graphicx}  
\usepackage{algorithm}
\usepackage{algpseudocode}
\usepackage{bm}
\usepackage{multirow}
\usepackage{amsmath}
\title{Learning Conceptual-Contextual Embeddings for Medical Text}
\author{Xiao Zhang\textsuperscript{\rm 1\thanks{Work done while visiting the University of Oregon}}\quad Dejing Dou\textsuperscript{\rm 3,\rm 4}\quad Ji Wu\textsuperscript{\rm 1,\rm 2}\\
		\textsuperscript{\rm 1}Department of Electronic Engineering, Tsinghua University\\
		\textsuperscript{\rm 2}Institute for Precision Medicine, Tsinghua University\\
		\textsuperscript{\rm 3}Department of Computer and Information Science, University of Oregon\\
		\textsuperscript{\rm 4}Baidu Research\\
		\texttt{xzhang19@mails.tsinghua.edu.cn}\\
		\texttt{dou@cs.uoregon.edu, doudejing@baidu.com}\\
		\texttt{wuji\_ee@mail.tsinghua.edu.cn}}

\begin{document}

\maketitle

\begin{abstract}
External knowledge is often useful for natural language understanding tasks. We introduce a contextual text representation model called Conceptual-Contextual (CC) embeddings, which incorporates structured knowledge into text representations. Unlike entity embedding methods, our approach encodes a knowledge graph into a context model. CC embeddings can be easily reused for a wide range of tasks in a similar fashion to pre-trained language models. Our model effectively encodes the huge UMLS database by leveraging semantic generalizability. Experiments on electronic health records (EHRs) and medical text processing benchmarks showed our model gives a major boost to the performance of supervised medical NLP tasks.
\end{abstract}

\section{Introduction}

External knowledge is often useful for language understanding tasks. Especially in specialized domains like medicine, it is unlikely to attain human-level performance in text understanding without referring to external domain knowledge. Ontologies and knowledge graphs are the most common forms of domain knowledge, but due to their structured nature, it is not straightforward to incorporate them with representation-based neural models.

Current approaches usually bridge text and knowledge graphs with retrieval. Triplets or entities are retrieved based on occurrences of the text tokens in the entity descriptions. After retrieval, triplets can be treated as text sequences and be provided to the model as an extra input \cite{Mihaylov2018}. Another method is to use the corresponding entity embeddings from a graph embedding model trained on knowledge graphs \cite{huang2019knowledge}. However one still needs to deal with the aligning issue between entity embeddings and text representations.

In this paper, we take a novel approach which takes external knowledge into the realm of text representation learning. Word embeddings models like skip-gram \cite{mikolov2013efficient} and contextual embedding models like BERT \cite{Devlin2018} have proved the crucial role of good text representations in NLP tasks. Our model aims to incorporate external knowledge into text representations, which makes it easy to apply external knowledge and makes it robust to variations of expression in text.

Our model, which we termed \textbf{C}onceptual-\textbf{C}ontextual (CC) Embeddings, is a contextual text representation model similar to BERT. Instead of providing general text representations, CC embeddings are specifically designed to be ``concept aware." The model is trained to recognize concept and entity names in text and produce representations of those concepts and entities. Knowledge from knowledge graphs is encoded in the representations, which can be easily utilized in NLP tasks. Like other contextual representation models, CC embedding model can be used to generate embeddings as features or fine-tuned for a supervised learning task.

The rest of the paper is organized as follows: we first formulate our approach and discuss why it is particularly relevant for the medical domain. Then we detail our model and the process of encoding a large knowledge graph into contextual representations. Finally we evaluate on several tasks to validate the effects of our CC embeddings.

\section{Methodology}

\subsection{Model}

\begin{figure}[h]
\centering		
\includegraphics[keepaspectratio=true, scale=.45]{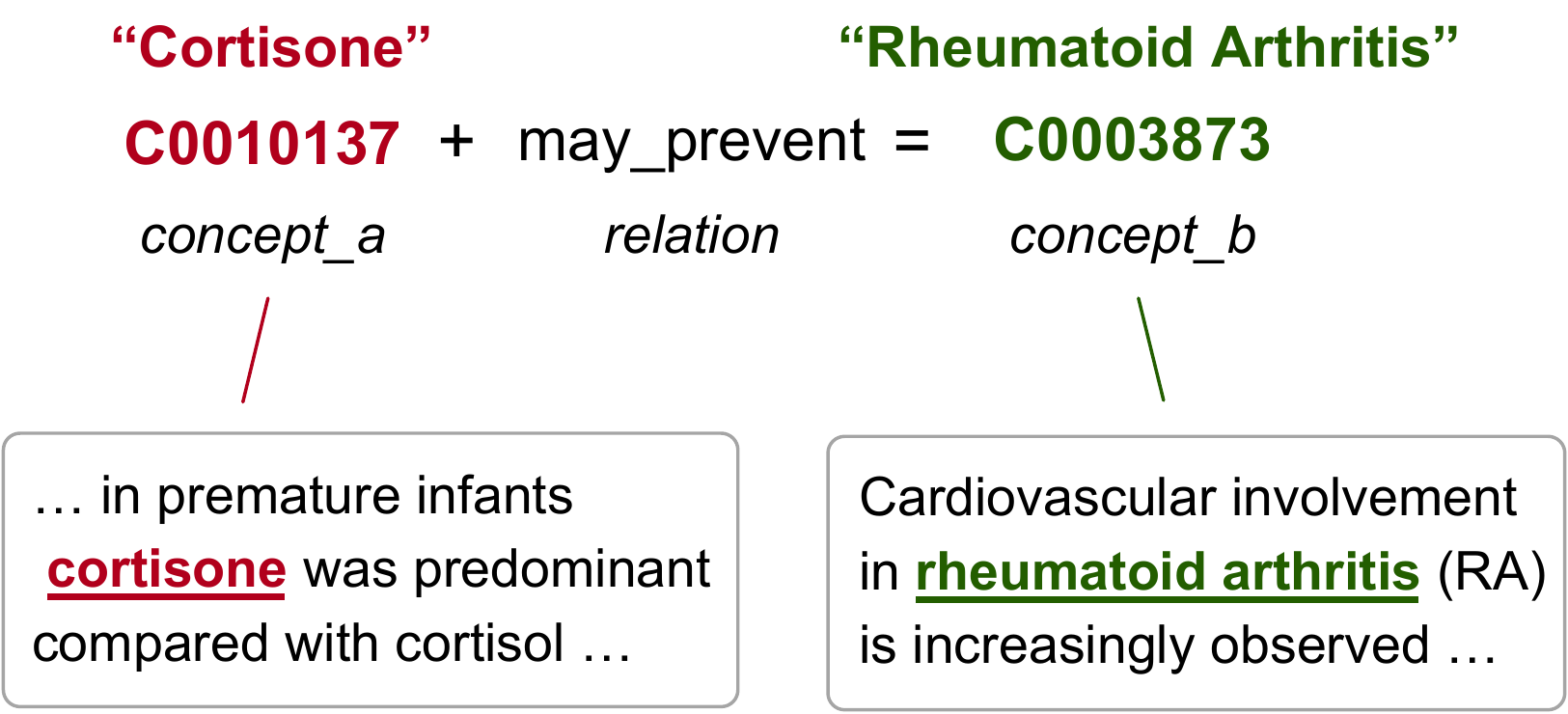}
\caption{Encoding concept mentions in text}
\label{fig:egs}		
\end{figure}

The core component of CC embedding model is an encoder which encodes structured knowledge. The encoder takes a written form of a concept as input, and outputs a vector representation of the concept. The idea is illustrated in Figure \ref{fig:egs}.

In this example, the encoder encodes a mention of a concept within a piece of text, and produces a concept embedding that satisfies a relationship defined in a knowledge graph:
\begin{multline*}
	Encode($``Cortisone"$) + may\_prevent \\ \approx Encode($``Rheumatoid Arthritis"$)
\end{multline*}

For simplicity we assume that the encoded concept embeddings and relation embeddings satisfy approximate translational relationship. Such a formulation is similar to TransE \cite{NIPS2013_5071}, but instead of learning entity embeddings, we would like to learn an encoder that can ``compose" the right concept representation from a mention found in text.

In this work we use a multi-layer bi-directional LSTM network as the encoder, similar to ELMo \cite{Peters2018}. Given an input sentence, it computes a representation vector at every word positions.  The only difference is that we adopt a knowledge graph embedding objective, rather than a language modeling objective.

To embed knowledge into it, the model is trained on a graph embedding task. During training the model is exposed to a large text corpus to learn to recognize and encode concepts in text. After training, the model absorbs structured knowledge in its parameters and is ready for reuse.

\subsection{Knowledge in medical KBs}

Medical text processing is still quite challenging despite recent success of deep learning in other modalities like medical image and sequential measurement data. The difficulty of incorporating a large amount of domain knowledge to understand text is definitely a reason. Therefore we are interested in getting an overall picture of medical domain knowledge from the perspective of NLP, and find out to what extent can representation models capture structured knowledge.

Another reason we are interested in the medical domain is that there exists a good collection of well structured domain knowledge, maintained in the form of multiple ontologies and knowledge bases (KBs). And more importantly, a large portion of the knowledge base entries has central attributes (like concept names, relation names) expressed in written language, rather than merely symbols and proper nouns. This makes text processing extremely relevant in utilizing the domain knowledge.

Diving into one of the medical KBs, one can summarize the typical structured information there into two categories:
\begin{itemize}
	\item \textbf{Language inferable (LI) knowledge}: these are triplets where the relation between two concepts can be at least partially inferred from the name of the concepts, e.g., 
	\item[\null]   \textit{Pulmonary Fibrosis} \enskip is\_sibling \enskip \textit{Cystic Disease of Lung}
	\item[\null]   The relation can be inferred as likely because ``pulmonary'' means ``relating to the lungs''.
	\item \textbf{Non-language inferable (Non-LI) knowledge}: facts that are independent of the meanings of the textual expression, for example:
	\item[\null]   \textit{Iodine 10 mg/ml Topical Solution} \enskip is\_a \enskip \textit{Ultracare Oral Product}
	\item[\null] In this example \textit{Ultracare} is a brand name, and it is impossible to infer whether the relation holds solely based on the above text. 
\end{itemize}

Research on knowledge bases usually do not make such distinctions and KB embedding models treat each concept as an individual entity. For text understanding, however, the first category deserves special attention because it represents \textbf{generalizable} knowledge. Such knowledge can be encoded in word representations or context representations, which can generalize to unseen expressions of concepts.

First, the knowledge can be generalized to different ways of writing the same concept. Medical concepts often have different names in different KBs, for example ``\textit{Enamel Dysplasia},'' ``\textit{Enamel Agenesis},'' and ``\textit{Enamel Hypoplasia}'' can refer to the same concept. Second, knowledge can also generalize from one concept to other concepts, such as  from ``\textit{Pulmonary hypertension}'' to ``\textit{Pulmonary Fibrosis}." As we will show in our experiments, exploiting such generalizations is a key to learning good medical concept embeddings.

Unlike entity embeddings, knowledge in text representations is generalizable and also directly available for neural NLP models and it can help text understanding in general. In the medical domain, applications that involve processing text such as doctor notes and electrical medical records could benefit from a representation model that incorporate generalizable domain knowledge.

\section{Related Work}

\textbf{KB embedding models.} In recent years a number of KB embedding models have been proposed that aim at learning entity embeddings on a knowledge graph \cite{cai2018comprehensive}. Some models make use of textual information in KBs to improve entity embeddings, like using textual descriptions of entities as complement to triplet modeling \cite{wangtrl,xiao2017ssp}, or jointly learning structure-based embeddings and description-based embeddings \cite{xie2016representation,ijcai2017-0183}. The latter approach learns an encoder which is similar to our work, but the encoder is only used to encode entity descriptions. These approaches are mainly concerned with KB representations rather than text processing. Using text also allows for inductive and zero-shot \cite{Yangrsl} entity representations, which is also a feature of our model.

\textbf{Word embedding models.} One way to incorporate external knowledge into text representations is by learning knowledge-enhanced word embeddings. Some use joint-objectives to train word embeddings to simultaneously satisfy co-occurrence relationships and external constraints, like in \cite{yu-dredze-2014-improving} and \cite{bianjiang}. Others rely on retrofitting, which fine-tunes word vectors in conventional word embeddings to reflect external knowledge \cite{faruqui-etal-2015-retrofitting,nguyen-etal-2016-integrating}. \cite{glavavs2018explicit} uses a technique called ``explicit retrofitting'' to learn a transformation which adds constraints to the embeddings. However, the external knowledge used in this line of work is mainly word-level lexical resources, like WordNet \cite{miller1995wordnet,liu-etal-2015-learning}, synonyms and antonyms \cite{nguyen-etal-2016-integrating,ono-etal-2015-word}. Integrating knowledge from a general knowledge graph is more difficult because entities and relations do not directly correspond to words.

\textbf{Concept embedding models.} In the NLP community sometimes concept embeddings are regarded as a form of phrase embeddings \cite{Mikolovdrw}, which can be learned by treating concepts as special words. One first annotate the concept mentions within a corpus, then use standard word embedding model to learn embeddings for those ``special words" \cite{vu-parker-2016-k,shalaby2018beyond}. In medical domain such method is widely explored with the help of automatic annotators and ontologies \cite{DeVine,finlayson2014building,choi2016learning}.  \cite{mencia-etal-2016-medical} expands the method by also using relationships found in structured text.

\textbf{Contextual representation models.} Recently contextual text representation models like ELMo \cite{Peters2018}, BERT \cite{Devlin2018} and OpenAI GPT \cite{radford2018improving,radford2019language} have pushed the state-of-the-art results of various NLP tasks. Language modeling on a giant corpus learns powerful representations, which provides huge benefits to supervised tasks, especially where labeled data is scarce. These models use sequential or attention networks to generate word representations in context. In the biomedical domain there is also BioBERT \cite{lee2019biobert}, a BERT model trained on PubMed articles that offers competitive results on medical text processing tasks. More recently some enhanced BERT models propose to mark entities in training, to make models aware of entities in text \cite{zhang-etal-2019-ernie,sun2019ernie}.

\textbf{Relationship to other knowledge-enhanced NLP models.} Some works have explored integrating knowledge representation into a specific task, like question answering \cite{hao-etal-2017-end,Mihaylov2018} and language inference \cite{chen-etal-2017-enhanced}. These models include network components to match entities and combine entity embeddings with input at inference time. The model design is usually specific to the task formulation, for example, a model designed on WebQuestions cannot naturally generalize to QA tasks where the answers are not restricted to be entities. By contrast, our approach encodes knowledge into a general text representation model, and no specific network structure is needed to leverage knowledge.

\begin{figure*}[t]
\label{fig:model}
\centering
\includegraphics[keepaspectratio=true, scale=.35]{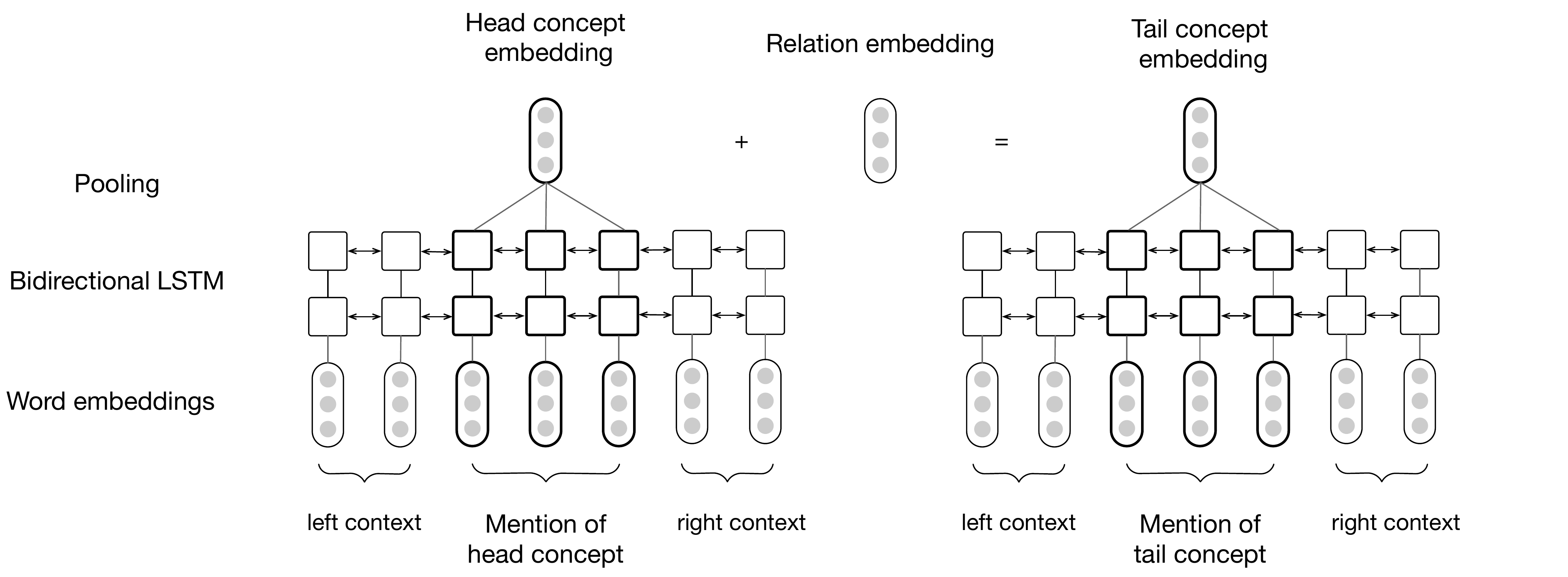}
\caption{Training CC embedding model to embed concepts}
\end{figure*}

\section{Conceptual-Contextual Embeddings}
In this section we detail the task used to train Conceptual-Contetxual embeddings and the training scheme, also performing evaluation within a knowledge graph for analyzing the effectiveness of training.
\subsection{Task}
To encode the knowledge into a text representation model, we use knowledge graph embedding task, like in \cite{NIPS2013_5071}. To show our approach is scalable to large knowledge graphs in the medical domain, we use the UMLS database \cite{bodenreider2004unified} for learning to encode medical concept embeddings.

\textbf{UMLS.} The Unified Medical Language System (UMLS) \cite{bodenreider2004unified} Metathesaurus is a large biomedical thesaurus containing concepts and relations from nearly 200 vocabularies (knowledge bases). A statistic of the database is given in Table \ref{table:umls_stat}. A concept in UMLS has one or more names associated with it (because different source vocabulary can name a concept differently). Relationships are given as triplets \textit{(head\_concept, relation, tail\_concept)}.

\begin{table}[h]
\caption{UMLS dataset statistics}
\label{table:umls_stat}
\begin{center}
\begin{tabular}{lr}
\textbf{Item} & \textbf{\#} \\
\hline
Entities (concepts) & 2,983,840 \\
Relations (general label) & 14 \\
Relations (additional label)& 936 \\
Train triplets & 23,029,716 \\
Test triplets & 8059 \\
\end{tabular}
\end{center}
\end{table}

\textbf{Concept names.} For each concept we take all the names associated with it. Among all the names, some are labeled as ``preferred name" by UMLS. We extract the first ``preferred name" as a primary name for the concept, and all the other as name variations.

\textbf{Relations.} Each kind of relationship in UMLS has a general label (REL) and an optional additional label (RELA). General labels describe the basic nature of the relationship (e.g., \textit{Broader, Narrower, Child of, Qualifier of}), while the additional labels explain the relationship more exactly (e.g., \textit{is\_a, branch\_of, component\_of}). We use additional labels as relationship labels whenever available, and use general labels when additional labels are absent.

All the triplets are extracted from the UMLS Metathesaurus Level 0 Subset and are split into a training set $T$ and a testing set. For each triplet in testing set, the triplet that describes the inverse relationship is removed from the training set (if found). We further removed concepts with non-latin characters in their names for more meaningful results in text-based models. 

\textbf{Context corpus.} Learning to recognize concepts in text requires ``seeing concept names in context." We prepared a corpus from PubMed citations and MIMIC-III critical care database \cite{Johnson2016}. Text is extracted from PubMed article abstracts and clinical notes in MIMIC-III health records. The corpus contains roughly 192 million sentences. We employ Apache Solr\texttrademark  \ to index the corpus, and use stemming normalization to increase recall for retrieval.

\subsection{Model and training}

\begin{algorithm}[t]
 \caption{Training CC embedding model}
\begin{algorithmic}[1]
\Require Training set of triplets $T=\{(h,r,t)\}$, relations $L$ and concept names $C=\{c_{1...n}\}$. Vocabulary $V$ and word embeddings $E$. Context corpus $S=\{s_{1...m}\}$
\Loop
\For{$(h,r,t) \in T$}
\State $(h',r,t') \leftarrow$ sample$(T, (h,r,t))$  \\ \qquad \qquad \qquad // sample a corrupted triplet
\For{$c \in \{h,t,h',t'\}$}
\State $c_{1...n} \leftarrow$ lookup$(C, c)$  \\ \qquad \qquad \qquad \quad// lookup concept names
\State $c_{1...m}^{ct} \leftarrow$ retrieve$(S, c_{1...n})$  \\ \qquad \qquad \qquad \quad// retrieve context sentences
\State $\bm{c}_{1...m}^{ct} \leftarrow$ LSTM$(E(c_{1...m}^{ct}))$
\State $\bm{c} \leftarrow$ selective\_pool$(\bm{c}_{1...m}^{ct})$
\State $\bm{c} \leftarrow \bm{c}/||\bm{c}||$
\EndFor
\State Update network w.r.t. \\ 
\qquad \qquad $\nabla[\gamma+d(\bm{h}+\bm{r},\bm{t})-d(\bm{h'}+\bm{r},\bm{t'})]_+$
\EndFor
\EndLoop \\
\Return Trained LSTM network (including modified word embeddings)
\end{algorithmic}
\end{algorithm}

The model is illustrated in Figure 2. The core of the model is a multi-layer bi-directional LSTM network. We make use of BioWordVec, a pre-trained biomedical domain word embedding from \cite{biosentvec} to embed text inputs.

Given a triplet $(h,r,t)$ in the training dataset $T$, we first lookup the name $h_{1...n}$ (with length $n$) of the head concept $h$: the primary name of the concept is used or, with probability $\alpha$, randomly replaced with one of its name variations. 

Next we use the name $h_{1...n}$ as keywords to retrieve sentences from the context corpus. We keep sentences with keyword occurrences lie adjacent to each other (forming phrases). A random sentence is selected from the top 10 ranked retrieval results as the context $h_{1...m}^{ct}$ for concept $h$.

The context sentence $h_{1...m}^{ct}$ (with length $m$) is then encoded by the LSTM network. The output sequence $\bm{c}_{1...m}^{ct}$ of the LSTM network is multiplied with a mask, which only keeps the output on the positions corresponding to the concept name in the sentence. The output is then max-pooled into a single vector $\bm{h}$ and normalized to unit length, which serves as a representation of the head concept $h$. The same is performed to generate a representation $\bm{t}$ of the tail concept.

Once the head and the tail concepts are encoded into vectors, we use vector addition in embedding space to model the relationship between the concepts. The formulation in this step is similar to TransE except that we use LSTM outputs in place of entity embeddings. For training the model, negative triplets $(h',r,t')$ are sampled by replacing the head or tail with a random concept, which are then processed by the model in the same fashion. The model is trained by minimizing a margin-based ranking loss:
\begin{equation}
	\mathcal{L} = \nabla[\gamma+d(\bm{h}+\bm{r},\bm{t})-d(\bm{h'}+\bm{r},\bm{t'})]_+
\end{equation}

In experiments we use 200-dimensional word embeddings and 2-layer bi-directional LSTM network with also 200 dimensions. In the ranking loss $\mathcal{L}$, Euclidean norm is used in distance function $d$ and margin $\gamma=0.1$. Vanilla stochastic gradient descent with learning rate $l=1.0$ is used to optimize the network. A total of 10 epochs is trained on 23 million training triplets.
Note that the hyper-parameter values are largely chosen heuristically and are not sufficiently tuned, due to efficiency reasons of the LSTM network and the size of the UMLS.

\begin{table*}[t]
\caption{Entity prediction results}
\label{table:entity_prediction}
\begin{center}

\begin{tabular}{lcccccccc}
\multicolumn{1}{c}{\multirow{2}{*}{\textbf{Model}}}  & \multicolumn{2}{c}{\textbf{Mean rank}} & \multicolumn{2}{c}{\textbf{Mean $\log$(rank)}} &\multicolumn{2}{c}{\textbf{Hits@10 (\%)}} &\multicolumn{2}{c}{\textbf{Hits@1 (\%)}}\\ 
& \it raw &\it filtered & \it raw &\it filtered & \it raw & \it filtered & \it raw & \it filtered \\
\hline
TransE         &213010&212298&2.90&2.90&16.7&16.7&3.3&3.3 \\
CC-DNN         &24441&23955&2.45&2.29&22.1&27.7&9.2&13.7 \\
CC-LSTM        &22888&22685&1.65&1.61&50.8&51.9&44.7&45.7 \\
CC-LSTM (DT) &43637&43518&\textbf{1.27}&\textbf{1.22}&\textbf{64.0}&\textbf{65.4}&\textbf{56.8}&\textbf{58.7}\\
\end{tabular}
\end{center}
\footnotesize{\qquad\quad\qquad\quad\quad *DT: discriminative training}
\end{table*}

\subsection{Discriminative Training}
To encode concept names into higher fidelity concept representations, the model needs to recognize  subtle differences between terms. We add a discriminative training step for this purpose. During training, when corrupted triplets are sampled, we sample concepts with names that are similar to the true concept instead of random sampling. This is done with probability $\beta=0.5$. For example, given concept ``\textit{Myeloid Leukemia}'' as a true tail, concept ``\textit{Lymphocytic Leukemia}'' would be more likely to be sampled as a corrupted tail under discriminative training. To avoid calculating the full similarity matrix between 3 million concepts, we take a crude but fast approximation: when sampling a negative concept $c'$, we randomly choose a word $w$ from the name $c_{1...n}$ of the true concept $c$, then randomly choose a concept $c'$ that also has $w$ in its name. The sampled negative concept $c'$ at least shares one common word in its name with the true concept.

This sampling step increases the difficulty of negative samples by making them more similar to the true triplets and thus more challenging for ranking. This forces the model to discriminate the semantic meaning of similar named concepts. To keep the model exposed to the whole set of possible concepts, there is still $1-\beta$ probability to sample from any concepts.

\subsection{Intrinsic Evaluation}
Before evaluating the learned representation on downstream tasks, we want to first analyze to what extent our model encodes structured knowledge in the UMLS, also validate the generalizability of the model.

We use the entity prediction task to measure the quality of the embeddings produced for concept names. Entity prediction is a standard task for evaluating entity embeddings, but here we only use it for analysis purposes rather than as a goal. For each triplet from the testing set we split from the UMLS, either the head or the tail is replaced with every concept in the UMLS. The true triplet is then ranked against corrupted triplets by the model. Results of ranking performance are shown in Table \ref{table:entity_prediction}. We follow common practice to report raw and filtered ranks.

TransE is listed in the table as a reference because we use the same translational formula to model relationships. CC-LSTM model performs surprisingly well on entity prediction, given that it is ranking among 3 million concepts. Especially for the Hits@1 metric, which is equivalent to making the ``correct" prediction. The best CC model makes the correct prediction more than half of the time, indicating its ability at fine-grained differentiation of concept semantics. 

We also include a mean $\log$(rank) metric, for a better representation of ``the average ranking position." When the number of ranking candidates is extremely large, one badly ranked example could make an otherwise good ``mean rank'' drop a lot, making the metric less intuitive. It can be seen from the mean $\log$(rank) column, roughly, the ranks of CC-LSTM model are generally of order $10^1$-$10^2$ and the ranks of TransE are generally of order $10^3$.

In place of the LSTM network, we experimented with using DNN to generate concept embeddings, but results are far inferior. Contextual information is important to correctly represent a concept based on its name. Discriminative training also substantially enhanced the performance of CC-LSTM model.

\begin{table}[h]
\caption{Performance on language-inferable and non-language-inferable knowledge}
\label{table:category}
\begin{center}
\begin{tabular}{lcc}
& \textbf{\# of examples} & \textbf{Hits@10 (\%)}\\
\hline
LI & 76 & 77.6 \\
Non-LI & 24 & 20.8 \\
Total & 100 & 64.0 \\
\end{tabular}
\end{center}
\end{table}

\begin{table*}[t]
\caption{Readmission prediction performance}
\label{table:readmission}
\begin{center}

\begin{tabular}{lccccccc}
\multicolumn{1}{c}{\textbf{Model}}  & \textbf{Acc} &\textbf{Pre-0} & \textbf{Pre-1}& \textbf{Re-0} & \textbf{Re-1} & \textbf{A.R.} & \textbf{A.P.} \\
\hline
\cite{lin2018analysis} &0.698&0.916&0.367&0.687&0.742&0.791&0.513 \\
LSTM          &0.840&	0.956&	0.366&	0.859&	0.704&	0.794&	0.600 \\
CC-LSTM       &0.848&	0.978&	0.321&	0.854&	0.786&	\textbf{0.804}&	0.613\\
\end{tabular}
\end{center}
\footnotesize{\qquad\qquad\qquad\quad\qquad\quad\ *Acc: Accuracy, Pre: Precision, Re: Recall, A.R: Area under ROC, A.P: area under PRC}
\end{table*}

\textbf{Break-down analysis} To measure the effect of semantic generalizability on model performance, and to understand the performance gap between the CC model and TransE, we first sampled 100 examples from the testing set, and labeled them to two categories: Language-inferable (LI) and Non-language-inferable (Non-LI), following our previous definition. Performance of the CC model on each category is shown in Table \ref{table:category}. First we observe that 3/4 of the triplets contain knowledge that can be inferred from text. This shows that in medical knowledge graphs, a majority of structured knowledge can potentially be carried by text representations. Making use of concept names can be difference-making in medical knowledge embeddings. In the case of the CC model, on LI type examples it gets to 77 percent hit at top10, while for Non-LI type the performance is much lower and is on par with the TransE model.

\begin{table}[H]
\caption{Error analysis by category}
\label{table:error}
\begin{center}
\begin{tabular}{lc}
\textbf{Category} & \textbf{Percentage} \\
\hline
Policy & 3.0 \\
Long name & 2.5 \\
UNK & 7.5 \\
SIB & 5.0 \\
Facts (Non-LI) & 14.5 \\
Other errors & 5.5 \\
\hline
Correct & 62.0 \\
\end{tabular}
\end{center}
\end{table}

On LI type knowledge the model is still quite far from perfect. We summarize the reasons for model failure in Table \ref{table:error}. After examining 200 examples in the testing set, we arrive at five common categories of difficult examples, which are:
\begin{itemize}
	\item Policy: this category is for triplets describing knowledge on medical policy or administration. These are not medical knowledge in the very strict sense, and the poor result could possibly be attributed to domain mismatch of pre-trained word embeddings.
	\item Long name: the name of one of the concepts in the triplet is longer than 10 words. Because we truncate long names to 10 words for faster training, some information is missing from the input.
	\item UNK: one of the concepts has more than half of out-of-vocabulary words in its name. This typically makes the concept indiscernible to the model.
	\item SIB: this category of triplets all have \textit{is\_sibling} relationship. The model seems to have some difficulty judging whether some closely related concepts are at the same level in the hierarchy.
	\item Facts (Non-LI): factual knowledge that is not inferable form text.
\end{itemize} 

These categories account for most of the errors of the CC model on the entity prediction task. Except for the Non-LI category, these errors are in principle resolvable with proper modifications to the model. Overall, the CC model captures language-inferable medical knowledge quite effectively, and next we will show it serves as a useful text representation.

\section{Downstream Applications}
As a contextual text representation model, the CC embedding model can be fine-tuned to various NLP tasks. By doing so the CC embeddings introduce concept awareness and external structured knowledge into the task model. We first present results on two real-world medical tasks then on another medical NLP benchmark task.

\subsection{MIMIC-III and Derived Datasets}
The MIMIC-III Critical Care Database \cite{Johnson2016,goldberger2000physiobank} is a large database of electronic health records (EHRs) of over 40,000 patients in Intensive Care Unit. Various kinds of numerical and report data is provided. In this study we are only concerned with textual data in EHRs. Specifically, we use the ``Discharge Summary" included in each ICU admission, which is a note written by doctors when the patient is discharged from ICU. Here is a snippet from one such note:

{\fontfamily{qcr}\selectfont \scriptsize
This is a 65 year old female with recent history of C. diff colitis (06') and recent mult abx use for UTI/PNA past couple months who presented after a syncopal episode in the setting of
diarrhea/dehydration ...}

Data pre-processing follows \cite{2017arXiv170307771H} and \cite{lin2018analysis}: after data screening there are 35,334 patients and 48,393 ICU stays. The patients are split into training (80\%), validation (10\%) and testing (10\%) sets with 5-fold cross validation.

In the following two tasks, we add a pooling and a linear layer on top of the CC-LSTM model to perform  classification. A plain LSTM classifier with identical structure is used as a baseline. All models use BioWordVec as word embeddings. We use early-stopping on validation set to select the best model. Reported results are averages over 5-fold splits.

\subsection{Readmission Prediction}

Unplanned ICU readmission rate is an important metric in hospital operation. Readmission prediction can help identify high-risk patients and reduce premature discharge \cite{10.1001/jama.2011.1515}. Our model predicts whether a patient is likely to be readmitted into ICU within 30 days, upon his/her discharge.

In Table \ref{table:readmission}, we present our model results and state-of-the-art result from \cite{lin2018analysis}. Lin et al. uses chart events, demographic information and diagnosis as input to a LSTM+CNN model. We only use written note text and none of the numerical and time-series information. The primary metric \textit{area under ROC} clearly shows that the CC model produces a performance boost over the baseline and surpassed state-of-the-art results.

\subsection{Mortality Prediction}
In this task we predict post ICU discharge mortality. Mortality prediction can help make better management and treatment decisions in costly ICU operations \cite{pirracchio2015mortality}. Table \ref{table:mortality} gives the prediction results of patient mortality within 30-day and 1-year after discharge. Note that like in the previous task, results from other works are listed mainly for reference rather than direct comparison, for these models use different information from EHR as input. Although not matching with state-of-the-art, the performance gain of the CC model over an LSTM model is consistent.

\begin{table}[t]
\caption{Post-discharge mortality prediction performance}
\label{table:mortality}
\begin{center}
\begin{tabular}{lcc}
\multirow{2}{*}{\textbf{Model}} & \textbf{30-day}  & \textbf{1-year}  \\
& \textbf{A.R.} & \textbf{A.R.} \\
\hline
\cite{ghassemi2014unfolding} & 0.80 & 0.77\\
\cite{ghassemi2014unfolding} & \multirow{2}{*}{0.82} & \multirow{2}{*}{0.81}\\
 \ (retrospective) \\
\cite{grnarova2016neural} & 0.858 & 0.853 \\
LSTM         &0.823&0.820 \\
CC-LSTM      &0.839&0.837 \\
\end{tabular}
\end{center}
\end{table}

\subsection{Medical Language Inference}
Natural language inference (NLI) is a task determining the entailment relationship between two pieces of text. We use the MedNLI dataset \cite{romanov2018lessons} to evaluate language inference in the medical domain. Original dataset contains 11232 sentence pairs for training and 1395 and 1422 pairs for development and testing. Results are listed in Table \ref{table:mednli}.

\begin{table}[h]
\caption{Performance on medical language inference}
\label{table:mednli}
\begin{center}
\begin{tabular}{lcc}
\multirow{2}{*}{\textbf{Model}} & \textbf{Dev} & \textbf{Test}  \\
& \textbf{Acc} & \textbf{Acc} \\
\hline
ESIM  & \multirow{2}{*}{74.4} & \multirow{2}{*}{73.1}\\
\cite{romanov2018lessons}& \\
ESIM (our implementation) & 74.8 & 71.3 \\
CC-ESIM         &\textbf{77.1}&\textbf{75.2} \\
\end{tabular}
\end{center}
\end{table}

We implemented the ESIM model \cite{2017arXiv170307771H} for NLI task, which consists of an LSTM encoder layer and an LSTM composition layer. CC-ESIM simply replaces the LSTM network in encoding layer with our trained CC-LSTM. The performance gain indicates the CC embeddings successfully introduces external knowledge into the model and benefits the task.

\section{Conclusion}
We have presented Conceptual-Contextual embeddings, a contextual text representation model which introduces structured external knowledge into text representations. The effectiveness of the model is validated on the medical domain, where domain knowledge is substantially associated with text understanding. Our work serves as a bridging perspective between knowledge graph representations and unsupervised text representation models. 

Future work include incorporating more powerful relationship models like TransR \cite{lin2015learning} into the CC embedding model. As our model only captures conceptual knowledge, combining CC embeddings with general representation models like BERT is also an interesting investigation. Under our formulation it is also straightforward to combine the two into a single model with multi-task learning, to further improve state-of-the-art text representation models.

\section{ Acknowledgments}
This research is partially supported by the National Key Research and Development Program of China (No.2018YFC0116800) and the NSF grant CNS-1747798 to the IUCRC Center for Big Learning.

\fontsize{9.5pt}{10.5pt} \selectfont
\bibliography{cc.bib}
\bibliographystyle{aaai}

\end{document}